\documentclass{article}

\usepackage[utf8]{inputenc}
\usepackage[english]{babel}
\usepackage{amsmath}
\usepackage{graphicx}
\usepackage{xcolor}
\usepackage[colorlinks=true, allcolors=blue]{hyperref}
\usepackage{subfigure}
\usepackage{array, multirow}
\usepackage[normalem]{ulem}

\title{FMAC: a Fair Fiducial Marker Accuracy Comparison Software}

\author{Guillaume J. Laurent and Patrick Sandoz}

\date{\small Université Marie et Louis Pasteur, CNRS, SupMicroTech-ENSMM\\
FEMTO-ST Institute, Besançon, F-25000, France.}

\begin{document}
\maketitle

\begin{abstract}
This paper presents a method for carrying fair comparisons of the accuracy of pose estimation using fiducial markers. These comparisons rely on large sets of high-fidelity synthetic images enabling deep exploration of the 6 degrees of freedom. A low-discrepancy sampling of the space allows to check the correlations between each degree of freedom and the pose errors by plotting the 36 pairs of combinations. The images are rendered using a physically based ray tracing code that has been specifically developed to use the standard calibration coefficients of any camera directly. The software reproduces image distortions, defocus and diffraction blur. Furthermore, sub-pixel sampling is applied to sharp edges to enhance the fidelity of the rendered image. After introducing the rendering algorithm and its experimental validation, the paper proposes a method for evaluating the pose accuracy. This method is applied to well-known markers, revealing their strengths and weaknesses for pose estimation. The code is open source and available on GitHub.
\end{abstract}

\section{Introduction}

Fiducial markers are patterns designed to be easily detected and localized within the field of view of an imaging system. Originally designed as simple circular or cross shapes, they now usually embeds a small code to be identified individually like tags. Fiducial markers have found numerous applications beyond augmented reality such as robot navigation and tracking in various fields (indoor robotics, construction robotics, drone docking, space, etc) \cite{alghamdi2025mobile}, 
camera and robot calibration \cite{atcheson2010caltag, filion2018robot, yin2024visionbased},
motion capture for virtual reality glasses \cite{borrego2016feasibility}, 
robot offline programming \cite{le2022solpen}, 
image-guided surgery \cite{wu2021closedloopa, lin2023modern}, structural analysis \cite{lei2020radial, cepon2024impactpose}, 
and even ethology \cite{mersch2013tracking}.  

Dozens of patterns have been proposed with different characteristics and merits. The metrics for appraising fiducial markers are numerous. The robustness of the detection is generally the most addressed one by testing its resistance to occlusion, defocus and motion blur, distortion, uneven illumination \cite{tsoy2022recommended, romero2021tracking}. 
This is usually done by estimating the rate of no detection, positive false, inter-marker confusion on image datasets. The range of detection, including the minimal size and the maximal orientation, is also an important feature to evaluate. The computational cost is of interest in case of embedded applications and fast tracking. 

When the markers are used as reference measuring systems, for example in construction and image-guided surgery, the accuracy of the pose estimation must be evaluated. 
The accuracy is usually expressed in terms of bias (difference between the expectation of the test results and the reference values). In the case of pose estimation, the bias has six components (three translations and three rotations) and depends on a lot of factors (marker size, camera used to record the images, distance between the camera and the maker, etc.).

Many papers investigate the evaluation of the pose accuracy that can be expected using a marker and a camera. The more convenient way is to track the marker position with a motion capture system but the resolution is then limited to a dozen of millimeters \cite{jurado-rodriguez2021design, kalaitzakis2021fiducial}. Some papers report using industrial robots as reference system \cite{yu2021topotag}, \cite{junior2024comparison}. However, industrial robots are precise but not really accurate. Using precision stages \cite{zhang2022deeptag} and precision robots \cite{ahmad2024motion} can improve results but these are still not metrology systems. The best solutions are to use a coordinate measuring machine or a laser tracker like in \cite{kedilioglu2021arucoe}, both being metrology systems. However, these are expensive and time-consuming to implement, thus limiting the number of poses that can be evaluated. Furthermore, it is challenging to reproduce the same conditions in different laboratories to conduct fair comparisons between markers.

\begin{figure}
\centering
\hfill
\subfigure[AprilTag \cite{olson2011apriltag, wang2016apriltag}]{\includegraphics[width=0.45\linewidth]{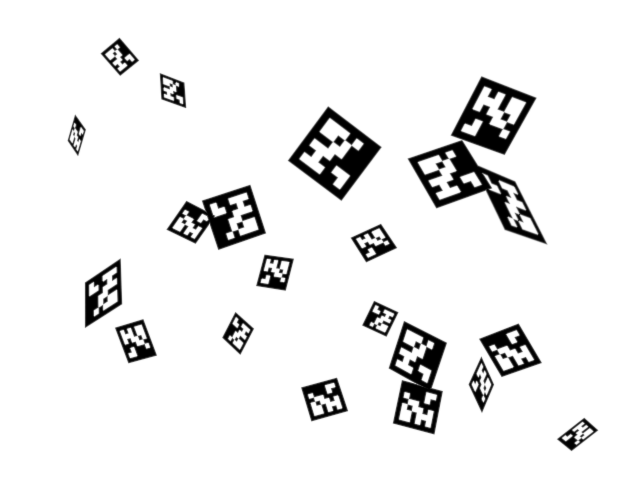}}
\hfill
\subfigure[STag \cite{benligiray2019stagb}]{\includegraphics[width=0.45\linewidth]{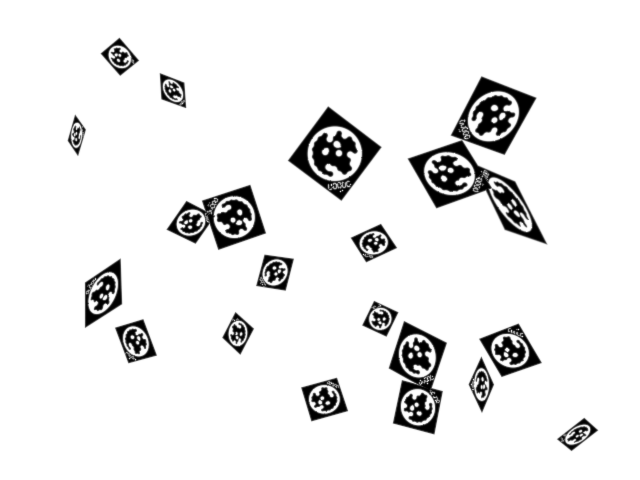}}
\hfill\,

\hfill
\subfigure[Topotag \cite{yu2021topotaga}]{\includegraphics[width=0.45\linewidth]{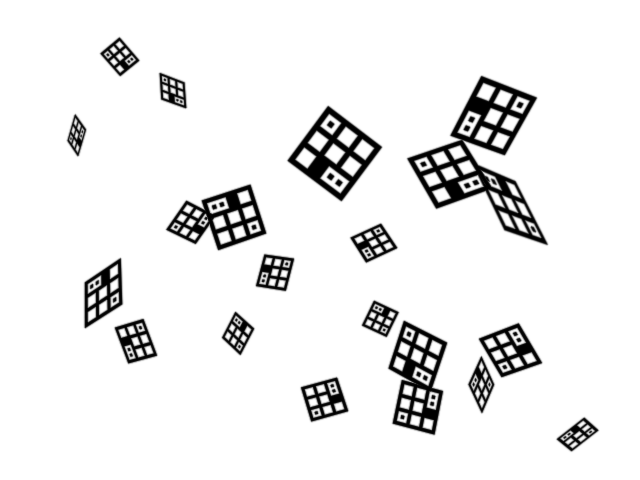}}
\hfill
\subfigure[ArUco \cite{garrido-jurado2014automatic}]{\includegraphics[width=0.45\linewidth]{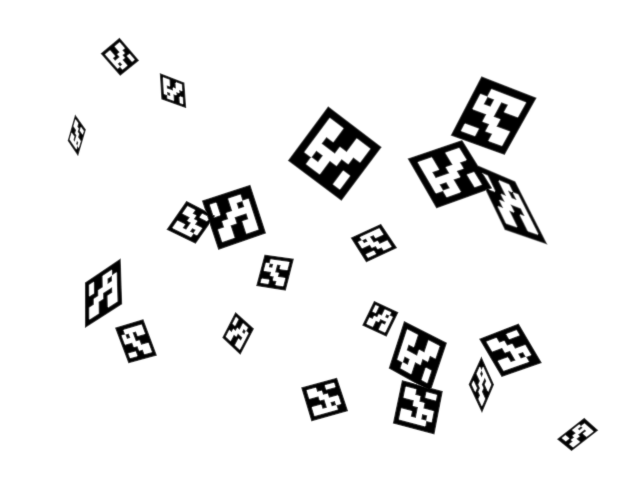}}
\hfill\,
\caption{\label{fig:snapshot}Synthetic images with different fiducial markers at the same poses (overlay of 21 images).}
\end{figure}

An alternative is to use synthetic images. Synthetic images have several advantages over experimental images. The ground truth can be generated without any errors nor perturbations that enables fair comparison between different markers by reproducing exactly the same poses and conditions. A large amount of data can be generated allowing an homogeneous sampling of the six degrees of freedom. However, the disadvantages are that synthetic images may be insufficiently realistic to reflect actual imaging systems. This is notably the case when using rasterization-based rendering, as used by robotic simulators like Gazebo \cite{lopez-ceron2022accuracy, shabalina2020artag}. Realistic images can be obtained using ray tracing. Ray tracing is based of the physics of light propagation and is able of rendering a variety of optical effects, such as shadows, optical and chromatic aberrations, scattering and caustics, but also the sampling of digital sensors. Several studies of fiducial markers use ray tracing as ground truth for accuracy evaluation \cite{bocco2021high, wu2014stable, rijlaarsdam2022novel,  olson2011apriltag}. 

In this paper, we present a physically-based ray tracing code{, FMAC, that was specifically designed to render images of fiducial markers. While ray tracing software, such as Blender, aims to render photorealistic images, FMAC focuses on rendering sharp edges accurately and on modeling defocus and diffraction blur. FMAC applies sub-pixel sampling to reproduce pixel gradients near the edges with a high fidelity. In addition, the code is built upon OpenCV and  directly uses intrinsic camera matrix and distortion coefficients, as obtained after calibration, along with physical characteristics of the image sensor (resolution, pixel pitch, bit depth) and settings of the lens (f-number, focus distance). FMAC can thus generate synthetic images of any fiducial marker at any poses as illustrated in Figure \ref{fig:snapshot}.

The purpose of this paper if to fulfill two needs. First, it provides a method to carry fair comparisons of the accuracy of several markers. Notably, FMAC allows a low-discrepancy sampling of the space to finely evaluate the performances along each degree of freedom and the resulting pose errors by plotting the 36 pairs of combinations. The second goal is to provide a tool for use in routine testing after camera calibration, to determine the level of accuracy that can be expected for any camera-marker couple. The code is open source and can be downloaded from GitHub at \url{https://github.com/vernierlib/fmac}.

\section{Physically Based Rendering}

Physically based rendering uses the principles of physics to model the interaction between light, optics and matter \cite{pharr2023physically}. The aim of physically based engines is usually to render photorealistic images. The objective is to produce an image that is indistinguishable from a photograph to the naked eye. In this paper, however, the goal is slightly different. All the fiducial marker detectors use the edges and the corners of a marker to compute its pose in the camera frame. The accuracy of the pose estimation depends directly on the position and shape of the edges and corners in the image. Small details, imperceptible to the human eye, must thus be rendered with a high fidelity and without aliasing to ensure reliable accuracy evaluation.

The proposed pipeline focuses on the rendering of sharp edges. Based on the thin lens model, it renders defocus aberrations with additional distortion corrections to reproduce geometric aberrations. Diffraction blur, which is very important at the pixel level, is also carefully modeled. The synthetic camera sensor is defined by its resolution, pixel pitch, bit depth and principal point. Finally, a local over-sampling is applied near edges to avoid aliasing. Table \ref{tab:parameters} gathers the constitutive parameters of the proposed rendering pipeline.

\begin{table}
    \centering
    \caption{Synthetic camera parameters with the two examples used for validation.}
    \medskip
    \small
    \renewcommand{\arraystretch}{1.5}
    \newcolumntype{M}[1]{>{\centering}m{#1}}
    \begin{tabular}{|M{2cm}|M{2cm}|M{2cm}|M{2cm}|M{2cm}|}
    \hline
    Model & Description & Notation & Logitech HD C270 & Canon EOS Rebel XS \tabularnewline \hline 
    \hline
    \multirow{4}{*}{Thin lens}
    & Focal length & $f$ & 4.47 mm & 26.49 mm \tabularnewline \cline{2-5}
    & Principal point & $c_x$, $c_y$ & 319.5 px, 239.5 px & 1407.5 px, 939.5 px\tabularnewline \cline{2-5} 
    & F-number & $N$ & 2.8 & 4.5 \tabularnewline \cline{2-5} 
    & Focus distance & $z_f$ & 150 mm & 700mm \tabularnewline \hline 
    \hline
    \multirow{3}{*}{Distortion} 
    & Radial coefficients & $k_1$, $k_2$, $k_3$, $k_4$, $k_5$, $k_6$ & -0.286, 0.057, 0.112, 0, 0, 0 & -0.125, 3.855, -40.371, 0, 0, 0 \tabularnewline \cline{2-5} 
    & Prism coefficients & $s_1$, $s_2$, $s_3$, $s_4$ & 0, 0, 0, 0 & 0, 0, 0, 0 \tabularnewline \cline{2-5}
    & Tangential coefficients & $p_1$, $p_2$ & 0, 0 & 0, 0 \tabularnewline \hline
    \hline
    \multirow{3}{*}{Sensor}
    & Resolution (image width and height) & $w$, $h$ & 640x480 & 2816x1880\tabularnewline \cline{2-5}
    & Pixel pitch & $\delta$ & 8.3 \textmu{}m & 5.7 \textmu{}m \tabularnewline \cline{2-5}
    & Bit depth & $8, 10, 12$ & 8 & 12 \tabularnewline \hline    
    \hline
    Diffraction
    & Wavelength of the light & $\lambda$ & 650 nm & 650 nm\tabularnewline \hline    
    \end{tabular}\label{tab:parameters}
\end{table}

\subsection{Thin lens model}

Most computer vision applications rely on the pinhole camera model, whose parameters are the focal length $f$ and the principal point $(c_x, c_y)$. Additional distortion coefficients are used to represent geometric aberrations.  However, the pinhole camera model only considers rays passing through a single point to reach the sensor, which is insufficient for reproducing defocus aberration. Real cameras have lens systems that focus light through a finite-sized aperture onto the sensor. The goal was thus to develop a code that uses the same camera parameters as those in standard libraries such as OpenCV, while extending the pinhole projection to the thin lens model. 

Under the thin lens approximation model, incident rays parallel to the optical axis pass at the focal point at a distance $f$ behind the lens. Any ray passing through the center of the lens is not deviated. The Gaussian lens equation relates the distances from the object to the lens $z_f$ and from the lens to its sharp image $z_s$: 
\begin{align}
    \frac{1}{z_f} + \frac{1}{z_s} = \frac{1}{f}
\end{align}
The focus distance $z_f$ can be adjusted on camera lenses and must be known in order to reproduce defocus aberration. If an object point does not lie at this distance, it is imaged as a disk on the sensor rather than as a single point. 
This disk is called the circle of confusion. Its diameter depends of the object point distance $z$ following:
\begin{align}
d_c = \left|\frac{d\cdot f\cdot (z - z_f)}{z\cdot(f+z_f)}\right|
\end{align}
where $d$ stands for the entrance pupil diameter (effective aperture), related to the f-number $N$ of the lens and to the focal length $f$ by:
\begin{align}
N = \frac{f}{d} 
\end{align}

As long as the circle of confusion remains smaller than the pixel pitch of the sensor, an object point will effectively appear to be in focus. 
FMAC uses this value to adjust the pixel sampling around the marker edges (see below).

\subsection{Distortion model}

Actual camera lenses are not thin lenses and produce geometric aberrations, primarily radial distortion and slight tangential distortion. The code uses the distortion model implemented in OpenCV, comprising six radial coefficients $k_1$,  $k_2$,  $k_3$,  $k_4$, $k_5$, and $k_6$, two tangential coefficients $p_1$ and $p_2$, and four thin prism coefficients $s_1$, $s_2$, $s_3$, and $s_4$. This model is defined by:
\begin{align}
\begin{bmatrix}
u' \\
v'
\end{bmatrix} = \begin{bmatrix}
u \frac{1 + k_1 r^2 + k_2 r^4 + k_3 r^6}{1 + k_4 r^2 + k_5 r^4 + k_6 r^6} + 2 p_1 u v + p_2(r^2 + 2 u^2) + s_1 r^2 + s_2 r^4 \\
v \frac{1 + k_1 r^2 + k_2 r^4 + k_3 r^6}{1 + k_4 r^2 + k_5 r^4 + k_6 r^6} + p_1 (r^2 + 2 v^2) + 2 p_2 u v + s_3 r^2 + s_4 r^4 \\
\end{bmatrix}
\end{align}
where $r$ is radial distance from the optical center, $r^2 = u^2 + v^2$.

Regular camera calibration procedures can be used to identify all distortion coefficients, as well as the focal length and the principal point.

\subsection{Ray tracing and sampling}

In ray tracing, an image is created by tracing the inverse path of a light ray from every sensor pixel through the camera lens until it intersects with an object. Multiple reflections can then be used to calculate the pixel intensity. The code assumes that the marker is perfectly illuminated without any specularity. Consequently, the visible color of the marker is simply the pixel value of its bitmap image.

The main difficulty in rendering a fiducial marker with a high fidelity originates from the aliasing of its edges. Most of fiducial markers are black-and-white binary bitmap images. This means that their edges are very sharp, creating abrupt transitions from black to white and vice versa. Rendering these high spatial frequencies therefore requires a high density of rays per pixel. However, the time taken to render an image is directly proportional to the number of computed rays.

To avoid aliasing without the computational expense of increasing the number of rays everywhere, adaptive sampling is used. First, a first coarse image is rendered with a single ray per pixel. The aim is to identify the regions of the image where over-sampling is required. Then, in a second step, pixels near edges are rendered again with numerous rays. 
To achieve the best results, the sampled points within a pixel must also have good spatial coverage without introducing bias. Several low-discrepancy samplers have been proposed, such as the Halton sampler and the Sobol sequence, the latter being faster to calculate. FMAC uses the open-source implementation of the Sobol sequence written by Leonhard Gruenschloss\footnote{Available at: \url{https://github.com/lgruen/sobol}}, that is based on the direction numbers \cite{joe2008constructing}. The resulting spatial sampling of a pixel is illustrated in Figure~\ref{fig:sampling}a.

Similarly, rendering defocus aberrations requires tracing many rays from each pixel in order to adequately sample the lens aperture. To this aim, a concentric mapping of the lens is used, as illustrated in Figure~\ref{fig:sampling}b. 
At the initial coarse rendering step, the size of the circle of confusion is used to determine the regions of the image where the lens must be sampled.

To determine the number of rays to be cast in the regions identified above, the code uses the quantization limit of pixel intensity. In a digital video camera, sensors are often 8-bit (i.e., 256 levels of gray), or 12-bit for professional cameras. Due to this quantization, it is unnecessary to cast more rays than the bit depth. To be sure to obtain the best results, the maximum number of samples per pixel is then the bit depth of the camera sensor, for example 256 rays per pixel for a 8-bit camera. 

\begin{figure}
\centering
\hfill
\subfigure[Pixel sampling with the Sobol sequence.]{\includegraphics[width=0.3\linewidth]{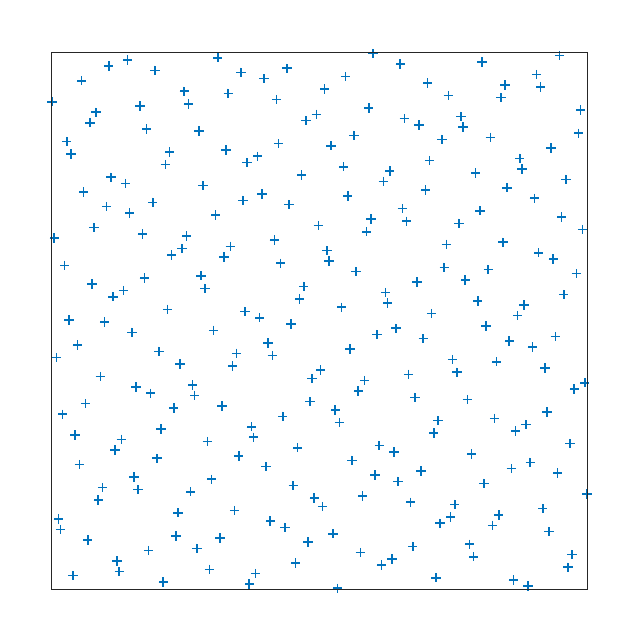}}
\hfill
\subfigure[Concentric mapping of the lens.]{\includegraphics[width=0.3\linewidth]{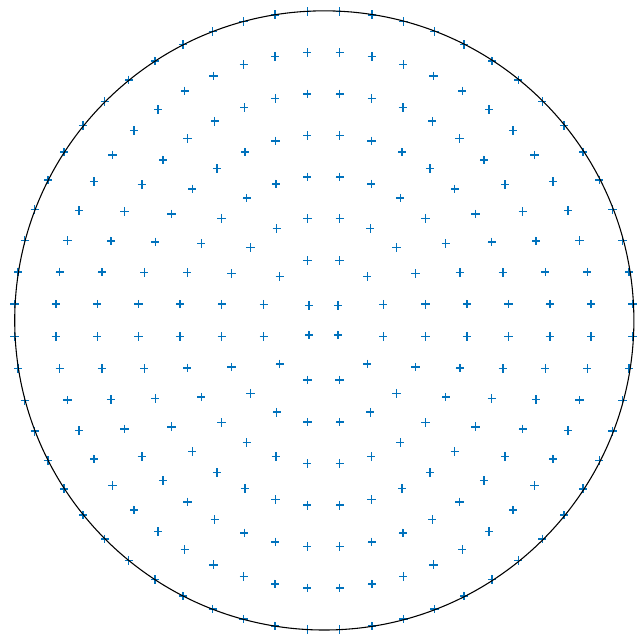}}
\hfill\,
\caption{\label{fig:sampling}Illustration of pixels and lens sampling.}
\end{figure}

\subsection{Diffraction blur}

Diffraction blur is often imperceptible in photorealistic renderings. At best, it is rendered using Gaussian filtering with a small kernel. However at the pixel level, it plays a significant role in the profile of a sharp edge. 
The continuous diffraction pattern of a circular aperture corresponds to a circular shape named Airy disk, defined as follows:
\begin{align}
g(u, v) = a \left[
                \frac{2\, J_1(\frac{\pi r}{r_a/r_z})}{\frac{\pi r}{r_a/r_z}}
            \right]^2       
\end{align}
where $a$ is the amplitude of the Airy function, 
$J_1$ is the first order Bessel function of the first kind, 
$r$ is the radial distance from the center of the kernel, 
$r_a$ the radius of the Airy disk (radius of the first zero) and 
$r_z = 1.2196698912665045$.

\medskip

For an optical lens, the radius of Airy disk $r_a$ is approximately $2\,r_z \lambda / d $, where $\lambda$ is the wavelength of the light and $d$ is the diameter of
the aperture. 
As this radius is of the same order of magnitude than the pixel size, the Airy pattern cannot be used directly as a discrete convolution kernel. 
To obtain a discrete diffraction kernel, the Airy pattern must first be convolved with a rectangular function $\Pi$ which is the size of a pixel. The resulting pattern is discretized at the pixel pitch $\delta$: 
\begin{align}
K(u,v) =(g * \Pi)(\delta u, \delta v)
\end{align}

The diffraction blur is added after the ray tracing phase by convoluting the image with the kernel $K$ that has been pre-calculated knowing the f-number and the wavelength of the light. 

\subsection{Gamma correction and pixel quantification}

The penultimate step in the pipeline involves applying gamma correction to the image intensity. The Rec. 709 transfer function \cite{bt1990basic} is used universally in almost all digital cameras:
\begin{align}
I'(x,y) = \begin{cases}
4.5 I(x,y) & \text{if } I(x,y)\leq0.018\\
1.099 I(x,y)^{0.45} - 0.099  & \text{else}
    \end{cases}
\end{align}
Finally, the corrected intensity is quantified according to the bit depth of the camera sensor (usually 8 bits or 12 bits).

\begin{figure}
\centering
\includegraphics[width=0.8\linewidth]{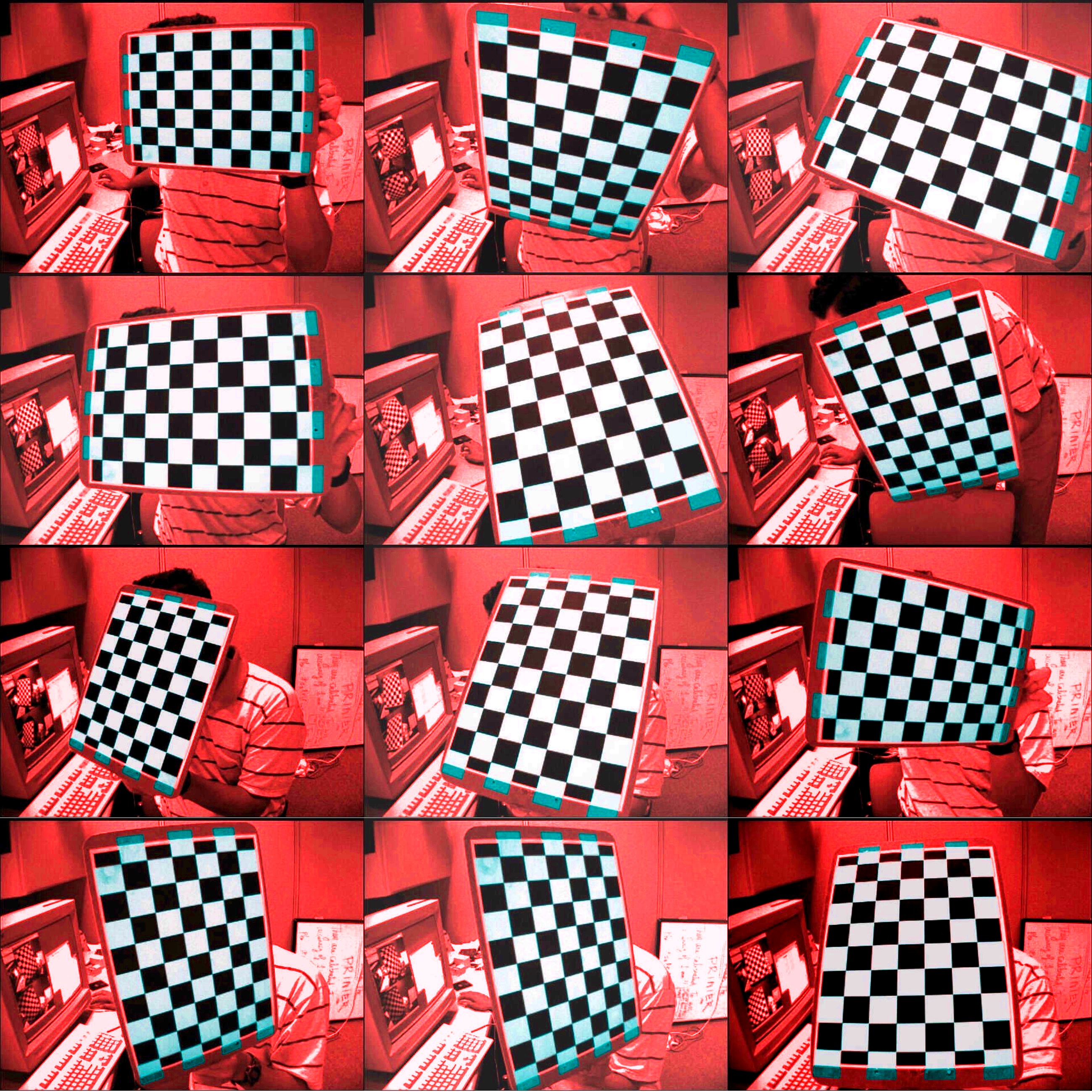}
\caption{\label{fig:projectionValidation}Overlay of actual and synthetic images from the OpenCV calibration dataset. The pixels colored in red indicate that the actual intensity is over the synthetic intensity. The pixels colored in cyan show the opposite errors. There is no difference between images if the pixels are white, black, or gray.}
\end{figure}

\begin{figure}
\centering
\hfill
\subfigure[Actual image]{\includegraphics[width=0.45\linewidth]{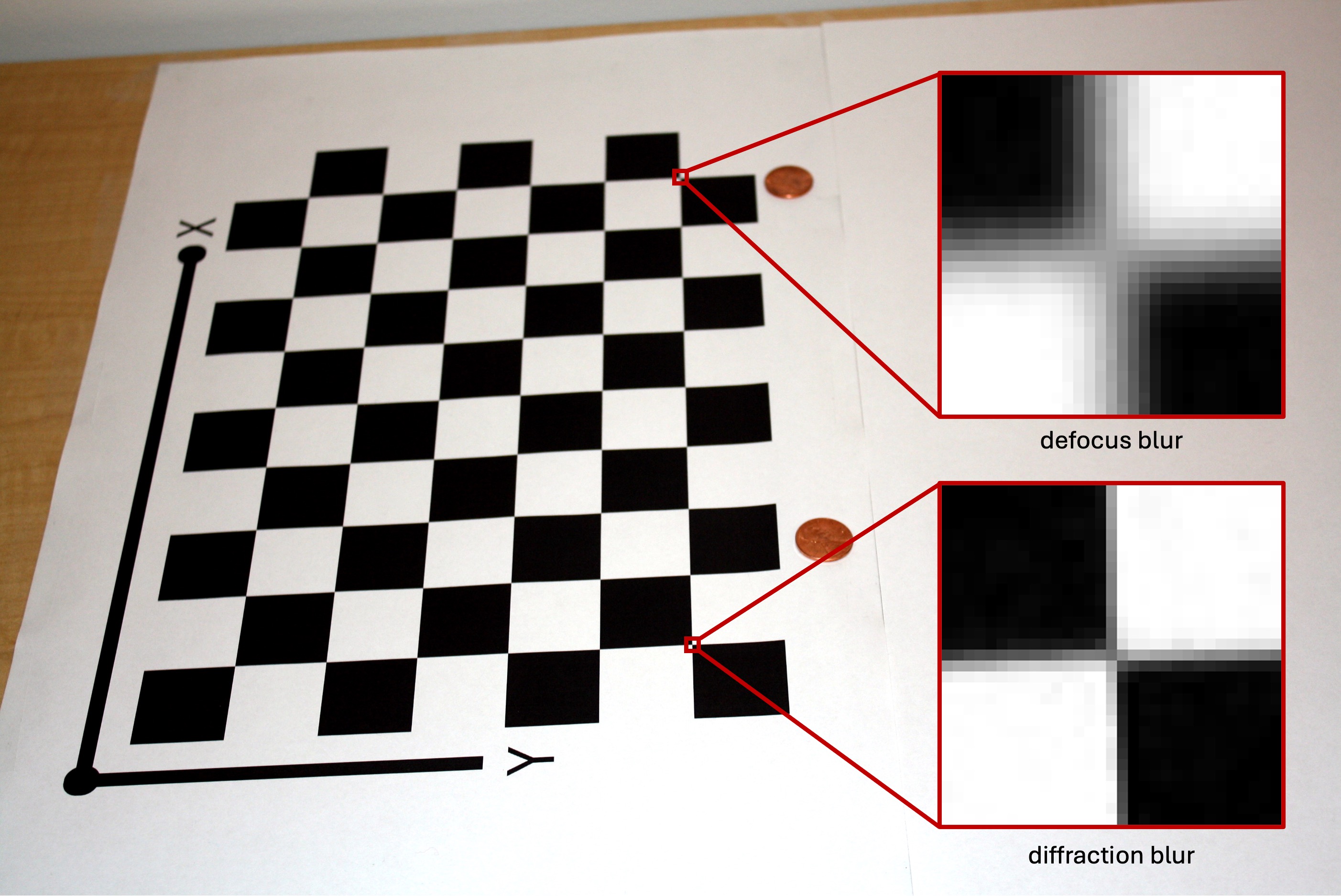}}
\hfill
\subfigure[Rendered image]{\includegraphics[width=0.45\linewidth]{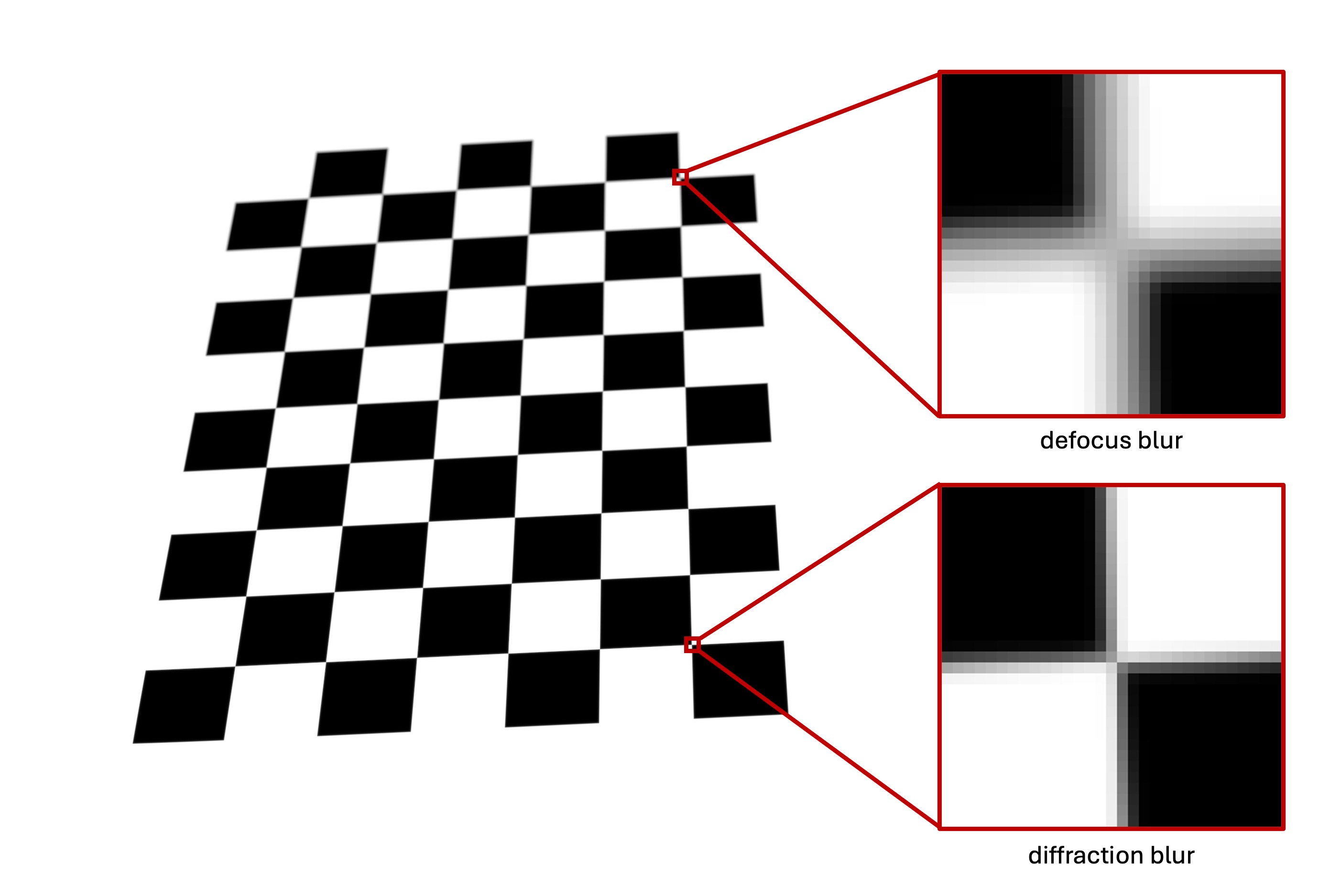}}
\hfill\,
\caption{\label{fig:depthValidation}Comparison of defocus aberrations on the Matlab's calibration dataset. The bottom zooms show details that are in focus. The top zooms correspond to points that are beyond the depth of field limit.}
\end{figure}

\section{Rendering Validation}

The quality of the rendered synthetic images was tested using two existing image sets. The first set comes from the OpenCV calibration examples. This set comprises 13 images of a chessboard captured with an 8-bit, 640 x 480 webcam such as the Logitech HD Webcam C270. The second set of images is part of MATLAB's calibration examples. This set contains nine images of a chessboard taken with a Canon EOS Digital Rebel XS with an 18–55 mm zoom lens. This camera produces high-quality 12-bit images with a resolution of 2816 x 1880. Information on the imaging conditions have been found in the EXIF data of the files or guessed on the image itself. The complete characteristics of both cameras are presented in Table~\ref{tab:parameters}.

The first image set was useful for verifying the projection model of the code. As the angle of view is large, the distortions are significant, enabling the thin lens and distortion models to be checked. The diffraction is relatively high due to the small size of the webcam's pupil. For the same reason, the depth of field is large, resulting in almost no defocus blur on the chessboard. Superimposing the rendered images on the actual images shows very good concordance, as illustrated in Figure \ref{fig:projectionValidation}.

The second image set was recorded with a higher f-number, resulting in significant defocus blur on some images. Figure \ref{fig:depthValidation} shows details of the actual and synthetic images of a chessboard. The bottom of both images is in focus, whereas the top is beyond the depth of field limit. The snapshot of the two corners shows a very similar gradient in both cases

\section{Fiducial Markers Accuracy Comparison}

The evaluation of fiducial marker accuracy is fraught with several challenges. In experimental systems, the ground truth is inherently prone to uncertainty. Furthermore, it is difficult to reproduce identical testing conditions for multiple markers.  Additionally, the process of capturing a large volume of images can be time-consuming. This section proposes a method for efficiently sampling the six degrees of freedom of the space using synthetic images. In order to enable fair comparison between various, large sets of synthetic images are generated for each marker but at identical poses. Then, the 36 combinations between each degree of freedom and each pose error are presented to assess potential correlations and trends of four markers. 

\subsection{Six degrees of freedom sampling}

The pose of a fiducial marker in the camera frame is described by at minimum six values, three translation distances and three rotation angles (seven values in case of using quaternions). Sampling a six dimensional space requires a lot of points. Moreover, naive sampling approach such as regular grids can introduce a bias in the results by favouring certain frequencies. 

In order to sample efficiently the space with minimal bias, we used the Halton sequence \cite{grunschloss2012enumerating}. The Halton sequence is known for its low-discrepancy, the idea that the points are distributed more evenly across the space, minimizing clustering and gaps. Halton's method generates points based on prime number bases (e.g., 2, 3, 5), which provides better coverage of the multidimensional space than purely random sampling. Like for Sobol sequences, the proposed code uses the open-source implementation of the Halton sequence written by Leonhard Gruenschloss\footnote{Available at: \url{https://github.com/lgruen/halton}} that supports both Faure-permutations and random digit permutations. 

We generated a cloud of 10\,000 images using the Logitech HD Webcam C270 parameters. The X and Y samples are restricted by the field of view taking the depth into account. 
The sampling of depth ranges from 500 to 1\,500 mm, ensuring that all the markers are in focus. The roll and pitch angle ranges are limited to -45 to 45 degrees. The yaw angle range is 360 degrees. 

\subsection{Pose estimation analysis}

This section compares the pose estimation errors of four well-known markers: ArUco \cite{garrido-jurado2014automatic}, AprilTag \cite{olson2011apriltag, wang2016apriltag}, STag \cite{benligiray2019stagb}, and TopoTag \cite{yu2021topotaga}. These markers have been chosen because each relies on a different way to achieve pose estimation. ArUco only uses the four corners of its outer square. AprilTag fits lines to outer edges gradient. STag uses the inner circular border to refine the estimated homography. TopoTag pose estimation takes advantage of all the inner marker vertices to increase the resolution. 

Each type of marker has been rendered in exactly the same poses, and its own detection code has been used to estimate the pose and compute the error against the ground truth. The size of the virtual markers are all 50x50 mm.

\begin{figure}
\centering
\includegraphics[width=\linewidth]{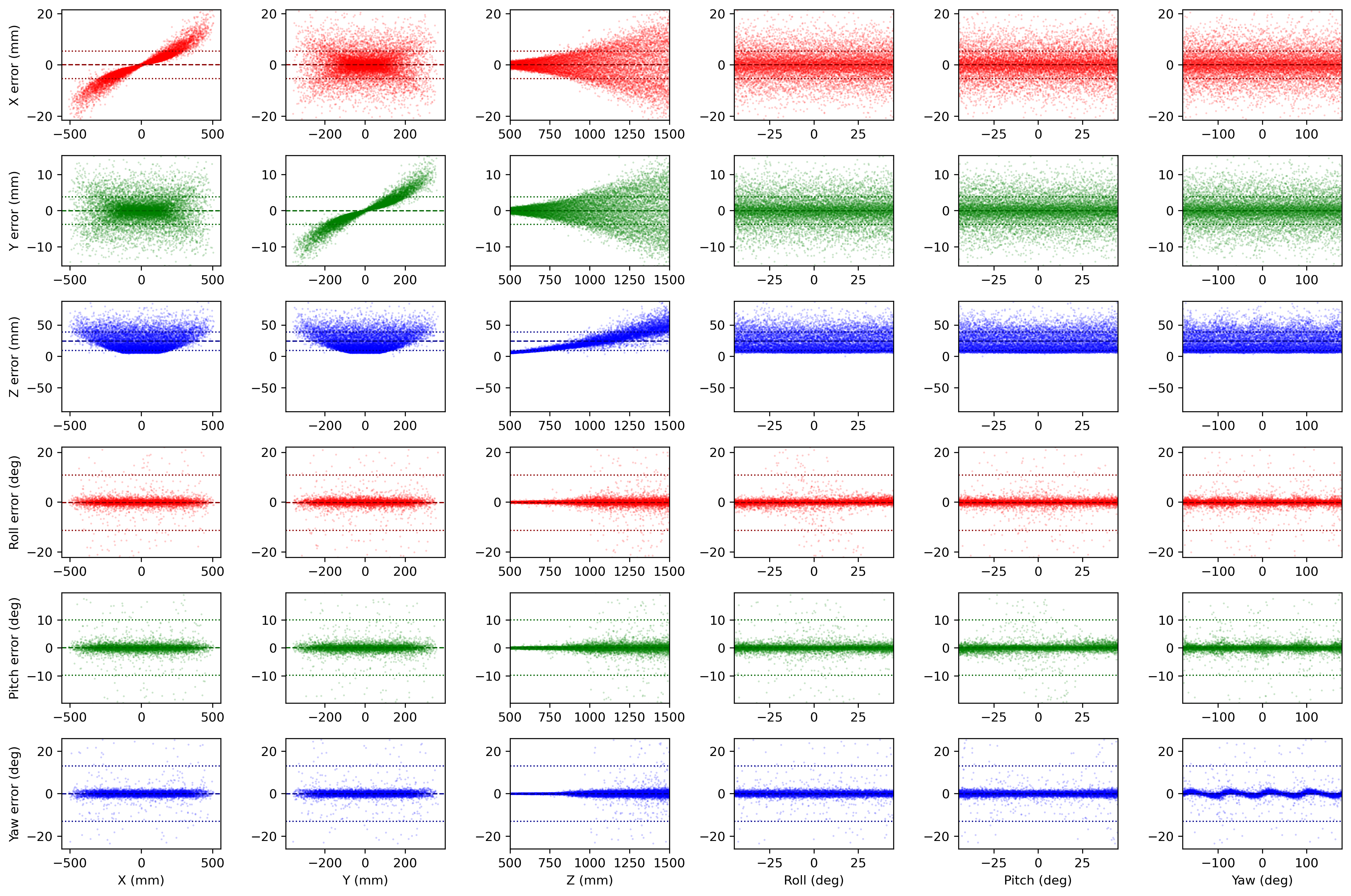}
\caption{\label{fig:correlations_aruco_logitech}Errors between the estimated poses of an ArUco marker and the actual pose for the six degrees of freedom. The images are rendered using the Logitech HD Webcam C270 parameters. The horizontal dashed lines represent the mean of the errors and the dotted lines the mean plus or minus the standard deviation.}
\end{figure}

\medskip

\paragraph{ArUco} The first marker tested is the ArUco, as implemented in version 4.13.0 of OpenCV. 
Figure \ref{fig:correlations_aruco_logitech} shows the 36 pairs of combinations between errors and values for the six degrees of freedom in the form of an intercorrelation graph matrix. 
This method allows to see the level of error for each value, as well as possible correlations with position/orientation in space. 

Throughout the whole measurement volume, the errors of pose estimation have a standard deviation of 5.4 mm in X and 3.8 mm in Y. These errors increase almost linearly as one moves away from the optical center. Errors in Z are more significant, reaching a standard deviation of 14.7 mm. They are always positive, which indicates that the detector overestimates the distance. The Z mean error is 24.2 mm and could be used to compensate the systematic overestimation. Rotation errors are fairly stable with a standard deviation in the order of a tens of degrees for all angles. The standard deviation is relatively large compared to the dispersion of the points because a small number of pose cause large errors, which can be explained by the detection ambiguity (less than 5 cases per thousand poses). 

A cyclical phenomenon appears clearly on the yaw angle with a period of approximately 90 degrees. This periodic error could indicate numerical errors in a trigonometric calculus. 
All errors increase significantly as the marker moves away from the camera and becomes smaller in the image. Overall, the pose estimation is much more accurate between 500 and 900 mm in depth. However, the rate of ArUco detection is 100\% on the whole image set.

\paragraph{AprilTag} Figure \ref{fig:correlations_apriltag_logitech} shows the same tests, this time using an AprilTag marker. Detection was carried using the AprilTag 3.4.5 release, which can be found on GitHub\footnote{\url{https://github.com/AprilRobotics/apriltag}}. 99.74\% of the markers are detected. A very few of small and angled markers are not detected. The errors in all degrees of freedom are significantly smaller. As with ArUco, errors in X an Y are linearly correlated with the distance from the optical axis. All the errors are also correlated with the depth, albeit to a lesser extent. AprilTag is particularly good at estimating the rotation angles. 
As with ArUco, the depth Z is always overestimated. The tests also revealed a systematic error in X and Y : the mean error is 0.62 mm in both directions. This error is equivalent to half a pixel, so we suspect that there is an unwanted constant offset in the code that should be removed in latter releases.

\begin{figure}
\centering
\includegraphics[width=\linewidth]{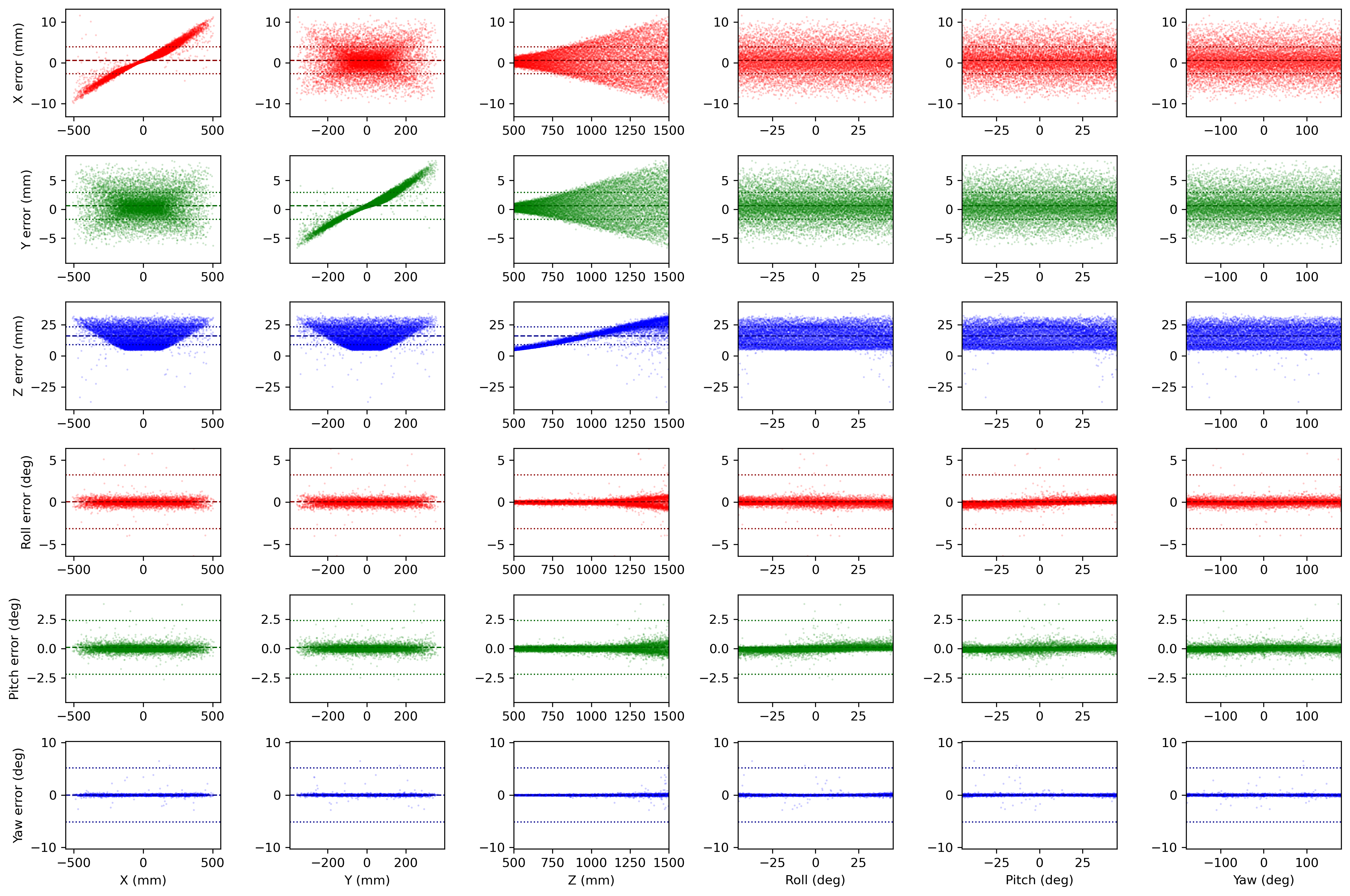}
\caption{\label{fig:correlations_apriltag_logitech}Errors between the estimated poses of an ApriTag marker and the actual pose for the six degrees of freedom. The images are rendered using the Logitech HD Webcam C270 parameters. The horizontal dashed lines represent the mean of the errors and the dotted lines the mean plus or minus the standard deviation.}
\end{figure}

\paragraph{STag} To evaluate the STag marker pose estimation, we used the Stoiber's fork of the original source code\footnote{\url{https://github.com/manfredstoiber/stag}}. Figure \ref{fig:correlations_stag_logitech} shows the detection results for the same set of poses. 

STag achieves a similar error level to AprilTag for the X, Y, and Z translations, but the standard deviation of angular errors is significantly higher than for ArUco and AprilTag. Unlike ArUco and AprilTag, STag does not systematically overestimate depth. However, the errors appear to be distributed around two levels, suggesting that the method uses two different estimation modes. 
The error in Z exhibits a periodic effect depending on the yaw angle. The detection rate of STag markers drops to 84.27\%. Markers smaller than 40 pixels or close to the image borders are poorly detected. 

\begin{figure}
\centering
\includegraphics[width=\linewidth]{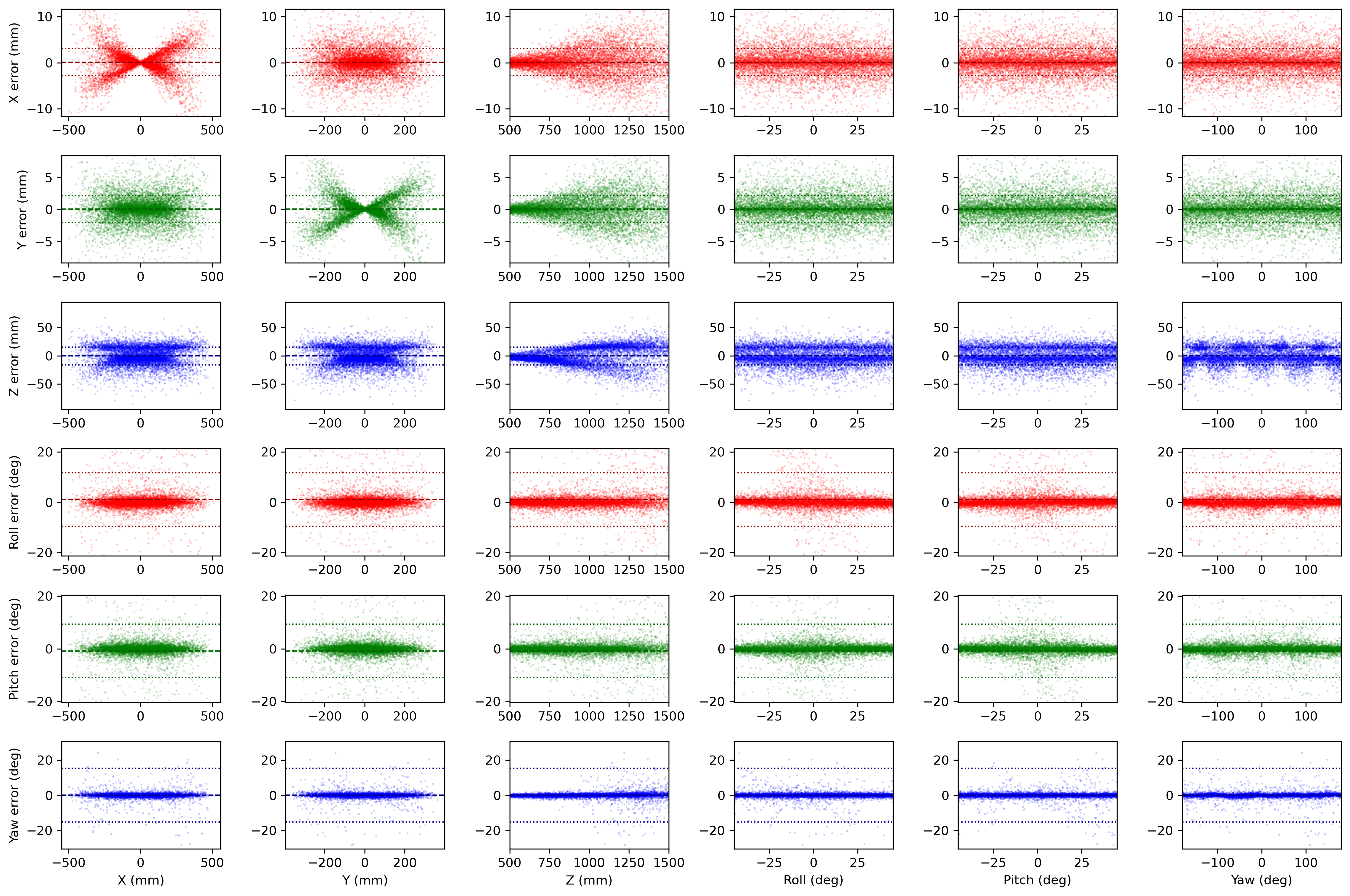}
\caption{\label{fig:correlations_stag_logitech}Errors between the estimated poses of a STag marker and the actual pose for the six degrees of freedom. The images are rendered using the Logitech HD Webcam C270 parameters. The horizontal dashed lines represent the mean of the errors and the dotted lines the mean plus or minus the standard deviation.}
\end{figure}

\paragraph{TopoTag} Finally, we tested the pose estimation of TopoTag using the binary code released on Github\footnote{\url{https://github.com/herohuyongtao/topotag}} (TopoTag is not open-source unlike the three other markers).

For the same pose set, the detection rate is only 57.65\%. TopoTags are correctly detected if their X and Y dimensions are greater than 50 pixels in the image corresponding to a distance of approximately 1000 mm. Beyond this depth, the detection rate drops dramatically, as can be seen in Figure \ref{fig:correlations_topotag_logitech}. 
On the other hand, when the marker is detected, the positioning errors are very low for all degrees of freedom. These errors remain remarkably constant throughout the space. We only notice a slight variation in X and Y errors when the yaw angle changes.

\begin{figure}
\centering
\includegraphics[width=\linewidth]{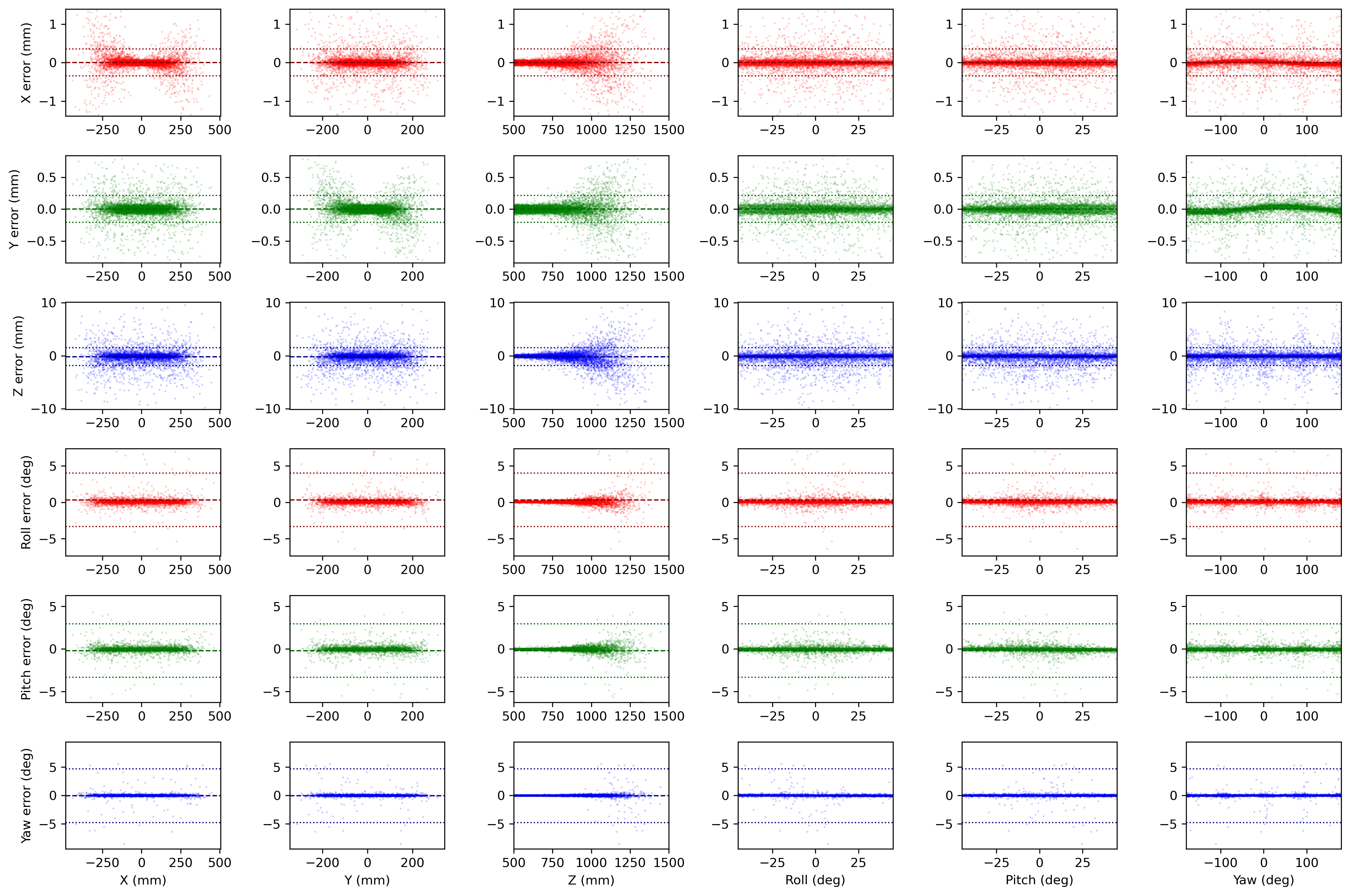}
\caption{\label{fig:correlations_topotag_logitech}Errors between the estimated poses of a TopoTag marker and the actual pose for the six degrees of freedom. The images are rendered using the Logitech HD Webcam C270 parameters. The horizontal dashed lines represent the mean of the errors and the dotted lines the mean plus or minus the standard deviation.}
\end{figure}

\subsection{Accuracy comparison}

\begin{figure}
\centering
\includegraphics[width=0.5\linewidth]{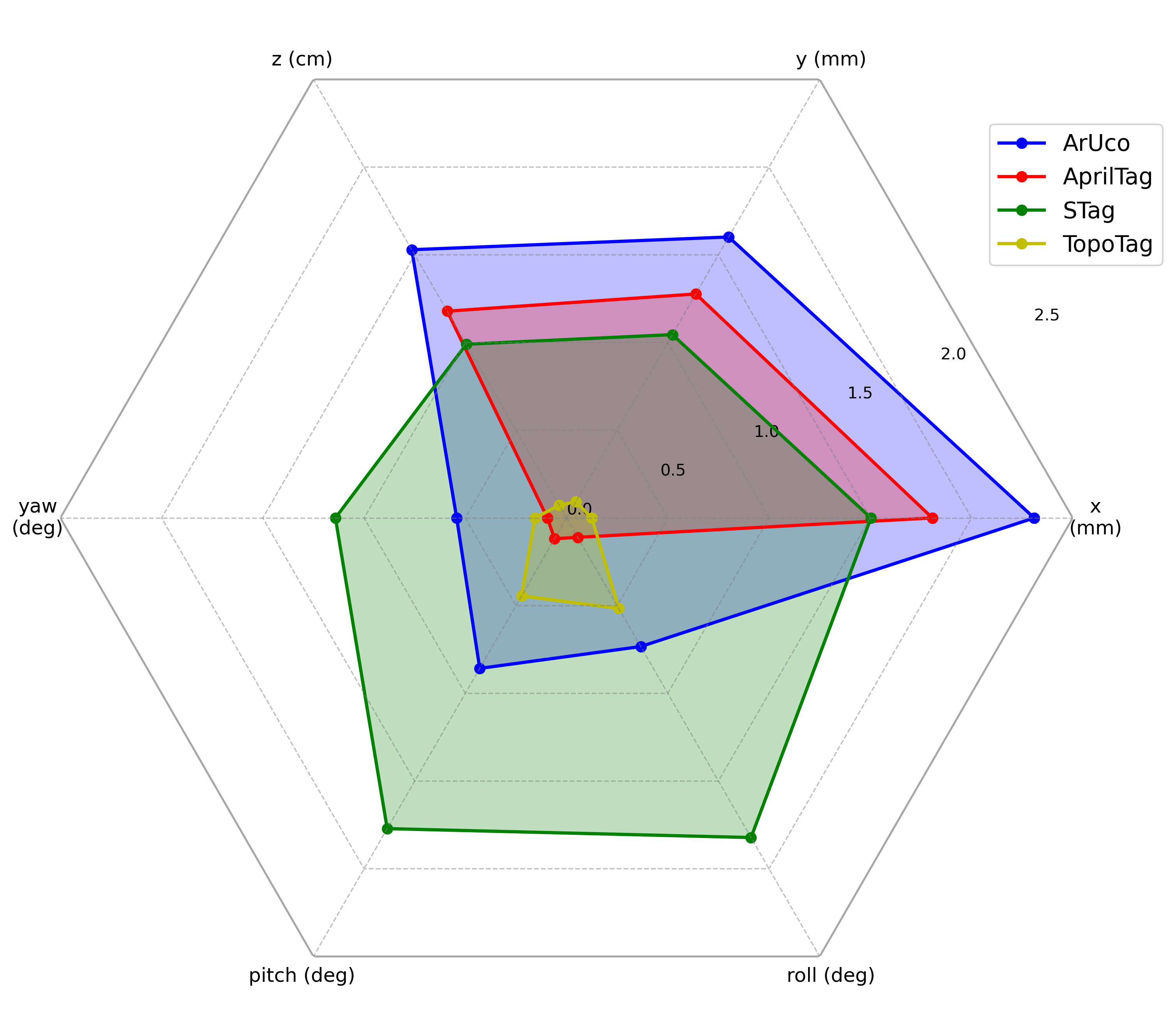}
\caption{\label{fig:accuracy_logitech}Comparison of the accuracy of four fiducial marker on the subset of 5625 poses that have been detected by all methods. The images are rendered using the Logitech HD Webcam C270 parameters.}
\end{figure}

According to the analysis, TopoTag appears to be the most accurate. However, a direct comparison would not be fair, since TopoTag markers are detected only for closest poses. For this reason, we calculate the accuracy with the only 5625 poses that were detected by the four methods. Moreover, the accuracy is defined as the closeness of agreement between the average value obtained from a large series of test results and an accepted reference value \cite{ISO5725}. Then, the accuracy corresponds to the average \emph{absolute} error and not to the standard deviation of the error as presented in previous figures. 
As our process is deterministic, one test per pose is sufficient. 
For instance, the positional accuracy for X is given by: 
\begin{align}
    A_x = \frac{1}{m}\sum_{k=1}^m |x_i - x_i^* |
\end{align}
where $m$ is the number of poses, $x_i$ the abscissa estimation of the ith  pose and $x_i^*$ its true value.

Figure \ref{fig:accuracy_logitech} presents the comparison of the accuracy of the four markers for the six degrees of freedom. It clearly shows that TopoTag is the most accurate method for estimating the three translations. The mean absolute errors are approximately 0.1 mm in the X and Y directions, and 0.7 mm in the Z direction. Nevertheless, AprilTag is superior for estimating the three angles with typical errors of around 0.1 degrees. It is important to note that TopoTag detection only works well at depths ranging from 500 to 1000 mm whereas AprilTag can detect smaller markers up to 1500 mm.

\section{Conclusion}

The methodology presented in this article aims to compare the accuracy of marker pose estimations using high-fidelity synthetic images. 
It allows to study the potential correlations between pose errors and the six degrees of freedom of the space. The graph matrices showing the 36 possible combinations are particularly useful for verifying the correct working of the detection algorithms.

This article focuses on accuracy estimation, but other marker metrics could be evaluated with the proposed FMAC software. For example, by using a camera with a smaller f-number, it would be possible to test robustness to defocus blur. Other disturbances such as occlusion, uneven illumination and low image dynamics could also be integrated into a standardized testing process to objectively qualify the performances of each type of marker.

\bibliographystyle{unsrt}
\bibliography{biblio}

\end{document}